# Vision based range and bearing algorithm for robot swarms

Hamid Majidi Balanji
Mechanical Engineering, Middle East Technical University
Ankara, Turkey
balanji.hamid@metu.edu.tr

Ali Emre Turgut
Mechanical Engineering, Middle East Technical University
Ankara, Turkey
aturgut@metu.edu.tr

*Abstract* — This paper presents a novel computer vision algorithm which proposed for the on-line range and bearing detection of the robot swarms. Results demonstrated the reliability of the proposed vision system such that it can be used for robot swarms applications.

*KeywordsS— Computer vision, robotic swarms, range, bearing*

## I. INTRODUCTION

Multi-robotic systems have inspired from the interaction behavior of social insects in nature. Robotic swarms is defined as the coordination of the large numbers of relatively simple robots in order to accomplish a common task [1]. In control theory of robot swarms, range and bearing information are used for flocking, collective transport and aggregation tasks [2]. Accordingly, It is really necessary for each agent inside a swarm to detect the neighbors' distance and position in order to avoid any collisions or do any control strategies inside swarms of robots. Knowing information about the range and bearing of the agents inside swarm is really crucial for some tasks such as: chain formation, self-assembly, coordinated movement, foraging and collision avoidance. Range can be defined as the distance from one robot to another and bearing defined as the horizontal angles between robots inside a swarm. Fig.1 (a) depicts these two terms schematically.

Nowadays, real-time measurements of the range and bearing can be considered as a challenging task. Researchers utilizes some expensive and hard-to-setup tools such as: Infrared, laser and radar for this purpose. Most of the researchers developed their swarm algorithms such as flocking and aggregations assuming that they have the range and bearing information in advance without any real physical measurements of them. Kelly and Keating calculated the range and bearing using Infrared sensors [3]. Turgut used short-range sensing system in order to measure the range and bearing of robots in close proximity [4]. Roberts et al., [5] applied 3D relative positioning sensor which was capable to calculate the range, bearing and elevation of the indoor UAVs. Accordingly, based on the mentioned materials, range and bearing measurements can be considered as an important approach in swarm intelligent. In this study, a novel and unique vision algorithm was proposed and developed such that it is able to measure range and bearing of the swarm agents by using a single camera and a passive landmark.

## II. METHOD

The hardware utilized for this study consisted of the Pololu Zumo chassis as the robot platform, raspberry pi model B+ as the real-time vision processing unit, Picamera as the vision module which can be integrated into raspberry pi and a testing area of 114 × 63 ×26 cm. Fig.2 shows the robot platforms and test setting configuration used in this experiment.

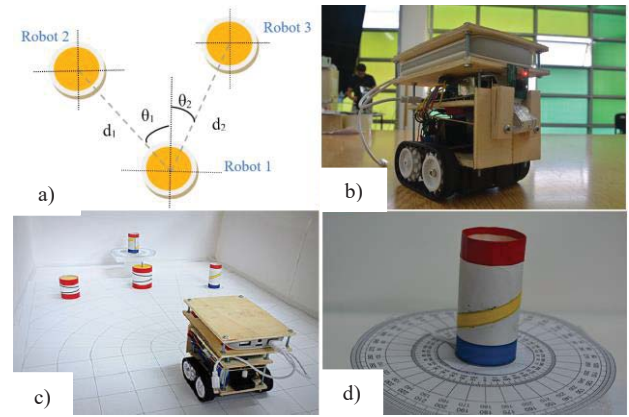

Figure 1. a) Range and bearing; b) robot platform; c) test setting area ; d) landmark

Landmark design was one of the main challenging part of this study. Several criteria set in the landmark design step. In other words, unlike other methods which utilized a special sensor mediums for the range and bearing detections, the proposed technique in this work was to design a landmark which contain information about the range and bearing gathered in a unified structure. Moreover, it must contain some geometrical and shape characteristics such that its information can be seen from different alignments and perspectives. For this purpose, a dozen of colors used in the landmark design such as purple, green, blue, yellow and etc. Additionally for the alignment problem, cylindrical patterns were selected since their shape and information on them can be viewed unchanged from every perspective. In the test setting light condition, the best results obtained from the red and blue colors. In other words, the red and blue colors were less sensitive to light variation and reflection when installed on a cylindrical surface. Finally, the target landmark designed as a cylindrical shape with the height 70mm and diameter 35mm and with red color ring at the top and blue color ring at the bottom according to the fig.1 (d).

## III. Experiment

The proposed vision algorithm was implemented in OpenCv with python libraries. Since all of the images were subjected to noises, therefore the first task was the noise suppressing procedure by using some noise removing filters in order to remove unwanted data. In this project, three popular filters such as Gaussian, median and bilateral filters applied. Among filters the best results obtained by the median filter of size 15. HSV color model was selected as the color space for the image processing since it is insensitive to light variations. The distance between the centers of the red and blue color regions used as a criterion for position detection of the swarms agents. Accordingly, before extracting the red and blue regions and calculating their centers, it was necessary to extract these regions from the other parts of the image. For this purpose, central momentums used as a powerful tool for the calculating of the centers of the red and blue regions. Fig.3 shows a simple schematic setting of the experiments.

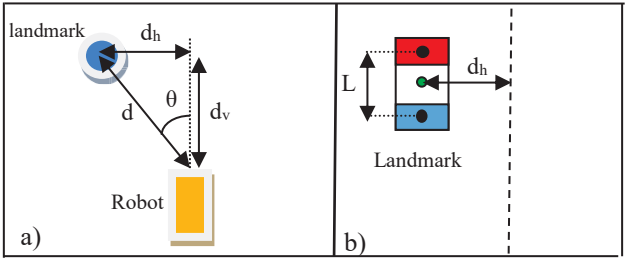

Fig. 3. Schematic view of the range and bearing calculation

In fig. 3, the distance between the red and blue regions, L, is calculated by the proposed vision algorithm in terms of pixel. In other words, L considered as an image feature which was used in this experiment for range and bearing calculations. The two parameters $d_v$ and $d_h$ are called the vertical and horizontal distances (cm) respectively which were manipulated automatically by the vision algorithm. The range and bearing were d and θ respectively. It is worth to mention that all of the mentioned distances in the fig. 3 are measured both manually and automatically in terms of centimeters and pixels respectively. Then a bunch of data sheets were compiled from the experiments.

## IV. Discussion and Results

After the statistical analyzing of the recorded data the relationship between the manipulated features L (pixel) and vertical distance $d_v$ (cm) calculated as:

$$d_v(cm) \propto \frac{k}{L(pixel)} \quad (1)$$

According to the fig. 3, for the range measurement, firstly it was required to measure the bearing angle. Thus, for every vertical distance, $d_v$, the corresponding the horizontal distance $d_h$ was calculated. After statistical analyzing of the recorded data, the pixel and centimeter based relationship between the $d_h$ as the distance from the center of the landmark into the center of the images calculated as:

$$d_h(cm) = kd_h(pixel) \quad (2)$$

After the calculation of the $d_h$ and $d_v$, the bearing angle calculated from the geometrical formulas as:

$$\theta = \text{atan}(\frac{d_h}{d_v}) \quad (3)$$

And finally, the range calculated as:

$$d = \frac{d_v}{\cos\theta} \quad (4)$$

## V. Conclusion and Results

In this study, a vision algorithm was proposed for range and bearing detection of the multi-robotic systems. Like every physical system in the technology, the proposed vision system suffered from some limitations. One of this deficiency in the proposed vision algorithm occurred when the robot was in the distance of the 28cm to 72cm of the landmark. In this distance t the equation (1) took the form of:

$$d_v(cm) = \frac{3500}{L(pixel)} \quad (5)$$

As it can be seen from the equation (1), L (pixel) and $d_v$ (cm) relation was related to each other inversely. In other words, as the robots got closer to the landmark, $d_v$ being decreased while L being increased. Another limit aroused from bearing angle detection which was able to detect the other agents in the bearing scope of the -25 to 25 degrees. After some statistical analyzes of the recorded data it was concluded that the error of the proposed vision system for the range and bearing detection calculated as 1.7% and 4.5 % respectively. Results in various light conditions, alignments and distances proved that the proposed vision system can be applied reliably for the robotic swarms' applications. Comparison of the proposed system with the other methods developed by the other researchers demonstrated that the proposed method in this work is simple, economically and computationally efficient because it utilized the passive landmarks such as print-out papers and a single Picamera, while in other researches most of them had applied some active, complex and expensive 3D sensors. A sample video of the real-world experiments can be found in [6]. The future work of this experiment has focused on the developing a vision-based system for the detection of the headings of the swarm agents.